\documentclass{article}
\usepackage{kr}
\usepackage[english]{babel}

\usepackage{amsmath}
\usepackage{graphicx}
\usepackage[colorlinks=true, allcolors=blue]{hyperref}
\usepackage{array, algorithm}
\usepackage{algorithmic}

\title{Random Client Selection on Contrastive Federated Learning for Tabular Data
}
\author{Achmad Ginanjar$^{a}$
\thanks{Indonesia Endowment Fund for Education Agency (LPDP)}
, Xue Li$^a$, Priyanka Singh$^a$, and Wen Hua$^b$  
\vspace{.3cm}\\
$^a$ School of Electrical Engineering and Computer Science\\
The University of Queensland,
Queensland, Australia\\
$^b$ Department of Computing, The Hong Kong Polytechnic University 
Hong Kong\\
}
\begin{document}
\maketitle

\begin{abstract}
Vertical Federated Learning (VFL) has revolutionised collaborative machine learning by enabling privacy-preserving model training across multiple parties. However, it remains vulnerable to information leakage during intermediate computation sharing. While Contrastive Federated Learning (CFL) was introduced to mitigate these privacy concerns through representation learning, it still faces challenges from gradient-based attacks. This paper presents a comprehensive experimental analysis of gradient-based attacks in CFL environments and evaluates random client selection as a defensive strategy. Through extensive experimentation, we demonstrate that random client selection proves particularly effective in defending against gradient attacks in the CFL network. Our findings provide valuable insights for implementing robust security measures in contrastive federated learning systems, contributing to the development of more secure collaborative learning frameworks.
\end{abstract}

\section{Introduction}
Vertical Federated Learning (VFL) \cite{Liu2024VFL} has emerged as a promising approach in collaborative machine learning. It allows multiple parties to jointly train models while maintaining data privacy through vertical partitioning of features \cite{liu2024vertical}. This paradigm has gained significant attention in privacy-sensitive domains such as healthcare and finance, where different organisations possess distinct feature sets of the same entities \cite{yang2019federated}.

Despite its potential to preserve data privacy, VFL faces inherent vulnerabilities related to information leakage during the intermediate computation sharing process. Research has shown that even partial information exchange can potentially expose sensitive data characteristics, compromising the fundamental privacy guarantees of the system \cite{Lyu2024}. These limitations have prompted researchers to seek more robust privacy-preserving solutions.

Contrastive Federated Learning (CFL) was introduced as an innovative approach to address these privacy concerns \cite{ginanjar2025contrastivefederatedlearningtabular}. By incorporating contrastive learning principles, CFL reduces the need for direct feature sharing while maintaining model performance through representation learning. This method has demonstrated promising results in minimising information leakage during the training process.

However, while CFL enhances privacy preservation in feature sharing, it does not fully address the broader spectrum of security threats in federated learning, particularly internal attacks. Among these, parameter-based attacks have emerged as a significant concern, where malicious participants can exploit parameter information to reconstruct private training data or compromise model integrity \cite{xia2023poisoning}. These attacks pose a substantial threat to the security of federated learning systems, potentially undermining their practical applications.

\begin{figure}
    \centering
    \includegraphics[width=1\linewidth]{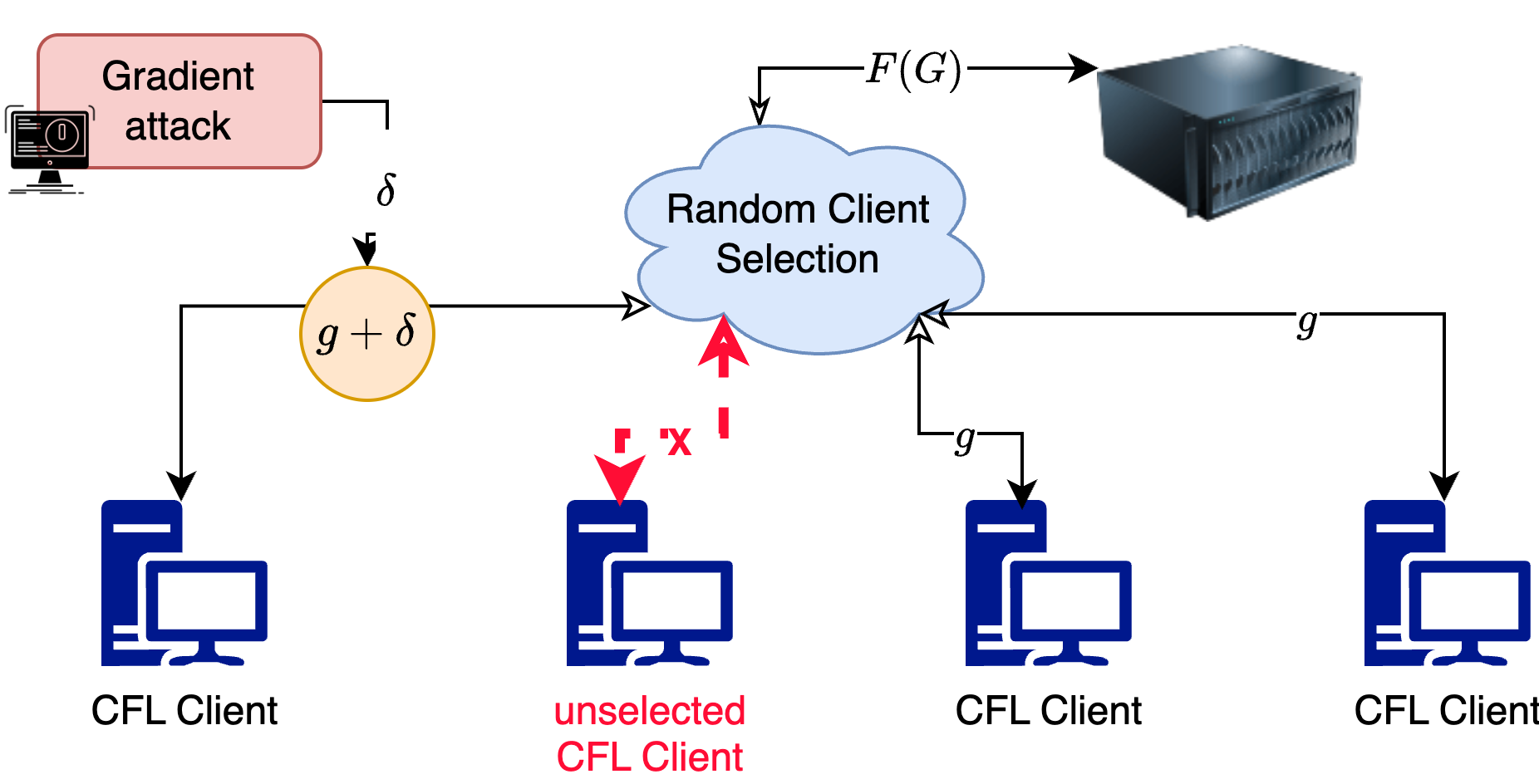}
    \caption{Random client selection within CFL network to defend poisoning gradient attack.}
    \label{fig:main}
\end{figure}

This paper presents three key contributions. First, it offers a systematic analysis of the impact of parameter-based attacks on contrastive federated learning (CFL) systems. Second, it demonstrates that employing a simple random client selection strategy serves as an effective defense mechanism against such attacks \cite{colosimo2024distance} , see Figure \ref{fig:main}. Lastly, it quantifies the effectiveness of this defence across 10 real-world datasets, encompassing a variety of attack scenarios.


Our experimental results show that random client selection can reduce attack success rates across all tested scenarios, while maintaining model performance. This finding is particularly significant for practical implementations, as it offers a computationally efficient defence mechanism  \cite{leiFu2023}  that can be immediately deployed in existing CFL systems. 

\section{Related Work}
This section reviews existing research relevant to contrastive federated learning, gradient attacks in federated learning, and client selection as a defence mechanism.

\subsection{Contrastive Federated Learning}
Contrastive learning \cite{Gutmann2010,Chen2020SimCLR} has been adapted to the federated setting to enhance privacy preservation.  Ginanjar \textit{et al}. \cite{ginanjar2025contrastivefederatedlearningtabular} introduce Contrastive Federated Learning (CFL) specifically designed for tabular data silos.  Their work focuses on reducing the reliance on direct feature sharing by leveraging representation learning, thereby mitigating potential information leakage.  This approach aims to achieve a balance between model performance and privacy protection in collaborative learning with vertically partitioned data scenarios.

\subsection{Model Attacks in Federated Learning}
Federated learning systems are prone to various attacks, notably those exploiting gradient and paramter information \cite{tolpegin2020data}.  Xia \textit{et al.} \cite{xia2023poisoning} provide a comprehensive survey of poisoning attacks in federated learning, detailing how malicious actors can manipulate parameters to compromise model integrity or infer sensitive training data.  A broader perspective on privacy and robustness in federated learning is offered by Lyu \textit{et al.} \cite{lyu2022privacy}.  They discuss a range of attacks, including gradient-based attacks, and present various defence mechanisms to counter these threats.  These studies highlight the critical need for robust security measures in federated learning.

\subsection{Client Selection in Federated Learning}
Lei Fu \textit{et al.} \cite{leiFu2023} work provides a broad explanation of federated learning client selection methods. They mention that although a random sample selection does not consider heterogeneity, this method is most likely selected. This is because beside other algorithm complexity \cite{wu2022node,luping2019cmfl,lai2021oort,zhou2022you} , This is because, besides other algorithm complexity \cite{wu2022node,luping2019cmfl,lai2021oort,zhou2022you} , these approaches rely on experiments to demonstrate their effectiveness. However, there is no guarantee of their performance in the real world.

In this study, we employ random client selection.

\section{Problem Formulation}
\subsection{Contrastive Federated Learning}
Let $D=\{(x_i,y_i)\}_{i=1}^N$ represent the training dataset, where $x_i$ denotes the feature vector and $y_i$ the corresponding label. In the vertical federated learning setting, the feature space is partitioned across $K$ parties, where each client $k$ holds a subset of features $x_i^k$. The CFL framework can be formalised as:
\begin{itemize}
    \item Client's objective is: \\ $f_c(\bar D( (\bar E :x;\omega^e);\omega^d)) \rightarrow x^d$
    \item Server's Objective is: \\ $ F(g) = \dfrac{1}{K} \sum_{k=1}^K (\omega^e,\omega^d) \rightarrow (\omega^{eG},\omega^{dG})$
\end{itemize}
\subsection{Model Attack}
The gradient-based poisoning attack in CFL can be formulated as follows:
Attack Objective: 
\begin{enumerate}
    \item A malicious party aims to reconstruct private information or compromise model performance by manipulating parameters: \\ $\min_{\delta} L_{attack}(\omega+\delta) $
where $\omega$ represents the true parameter and $\delta$ the poisoning attack.
    \item Attack constraints: \begin{itemize}
        \item Parameter manipulation must remain within bounds to avoid detection: $||\delta|| \leq \epsilon$
        \item The poisoned gradients should maintain statistical similarity to legitimate updates: \\ $||stats(\omega+\delta) - stats(\omega)|| \leq \tau$
    \end{itemize}

\end{enumerate}

Our study applies model scaling attacks. Applies
$\omega + \delta = \omega \cdot \alpha \text{ where } \alpha \in R^+ $ and $\alpha$ is the poison level.
\subsection{Random Client Selection}
The random client selection mechanism can be formalised as:
\begin{enumerate}
    \item Selection Process \\At each training round $t$, a subset of clients $S_t$ is randomly selected from the total client pool: \\ $S_t \subset {1,...,K},|S_t=m|$ \\ where $m$  is the number of clients selected per round.
    \item Selection probability \\Each client $k$ has an equal probability of being selected: $P(k \epsilon S_t)= \frac{m}{K}$
    \item Defence objective \\ The random selection aims to minimise the attack success probability: $ \min_{S_t}  P(Attack_{success} \| S_t$

\end{enumerate}

\subsection{Combined Defence Framework}
The overall defence framework integrates these components:
    \begin{itemize}
        \item Select clients: $S_t \sim \text{Uniform}(K,m)$
        \item Update local models: $\omega_k^{t+1}=\omega_k^t \eta \bigtriangledown L_{cont} (\omega_k^t) $
        \item Aggregate updates: $\omega^{t+1} = \frac{1}{|S|} \sum_{k \in S_t} \omega_k^{t+1} $
    \end{itemize}
\subsection{Theoretical Analysis}
Consider $\omega_t$ is the model parameter at iteration $t$, $\eta$ learning rate, and $\mu$ strong convexity parameter.
Given the poisoning probability $p_c$ and selection ratio $r_l$, the attack success probability is bounded by $P( \text{attack}) \leq p_c \cdot r_l$. 
Under random selection ratio $r_l$, the expected convergence satisfies:
$ E[||\omega_t - \omega*||^2]  \leq (1 - \eta \mu r_l)^t ||\omega_0 -\omega *||^2$. 

\section{Experiments}
The experiments in this study use ten datasets: Adult \cite{adult_2}, Helena \cite{helenaJannis}, Jannis \cite{helenaJannis}, Higgs Small \cite{higgssmall}, Aloi \cite{aloi},  Epsilon \cite{epsilon2008}, Cover Type \cite{covtype}, California Housing \cite{ca}, Year \cite{year}, Yahoo \cite{chapelle2011yahoo}, and Microsoft \cite{microsoft}. 

We perform an extensive study to challenge CFL. We use the model scaling attack proposed by Cao \textit{et al.} \cite{Cao2023FedCover}. We intentionally do not apply any defence, such as Byzantine \cite{s02021Byzantine} and other security enhancements \cite{lyu2022privacy} to the CFL network. This was done to test the robustness of CFL. This study covers three different parameters as shown in Table \ref{tab:settings}.
\begin{table}
    \centering
    \small
    \begin{tabular}{|p{0.17\textwidth}|p{0.051\textwidth}|p{0.16\textwidth}|} 
        \hline 
         Strategy&  Not& Information\\ \hline 
         Client poisoning number&  $p_c$& number of clients being poisoned.\\ \hline  
         Scaling poisoning level&  $p_l$& $\omega \cdot \alpha \text{ where } \alpha = p_l$\\  \hline
         Random client level&  $r_c$& number of clients skipped during FL.\\ \hline 
    \end{tabular}
    \caption{Poisoning attack settings where $\{k,p_c,p_l,r_l\} \in Z $. The 'Not' is the notation.}
    \label{tab:settings}
\end{table}

The comprehensive settings for our experiments are presented in Table \ref{tab:details_settings}. Across all datasets, we executed a total of 19 distinct experiments. 

To illustrate, consider an experiment that is assigned specific parameters: $\{n:8,p_c:0.2,p_l:0.1,r_c:0.2\}$, this indicates that the experiment is composed of 8 clients operating within a federated learning network. Among these clients, a designated subset of clients is classified as adversarial, each exhibiting a specific level of $(\omega\cdot 0.1 )$ data poisoning designed to test the robustness of the learning model. The global server is responsible for aggregating the model parameters, collecting data only from a random $int (8 \cdot 0.2 ) = 2$) of the clients. This methodology aims to simulate real-world scenarios where malicious clients may attempt to disrupt the learning process, allowing us to assess the robustness of  CFL under various attack conditions. The details of the experiments are provided in Algorithm \ref{alg:CFL-ADV}. 

\begin{table}
    \centering
    \small
    \caption{Parameters used for the experiments.}
    \begin{tabular}{|c|c|} \hline 
         Notation& values\\ \hline 
         $p_c$& $[0,0.2,0.5]$\\ \hline 
         $p_l$& $[0.1,0.5,2]$\\ \hline 
 $r_c$&$[1,0.2,0.8]$\\ \hline
    \end{tabular}
    
    \label{tab:details_settings}
\end{table}

\begin{algorithm}[]
\label{alg:CFL-ADV}
\caption{CFL with Potential Model Poisoning}
\begin{algorithmic}[1]
\STATE \textbf{Initialise:} 
\STATE \quad Global model \(F_G\)
\STATE \quad Set of clients \(C = \{C_1, ..., C_n\}\)
\STATE \quad Poison clients \(P_c \subset C\) randomly selected
\FOR{each epoch \(e = 1, ..., E\)}
    \FOR{each batch \(b\)}
        \FOR{each client \(c_i \in C\)}
            \STATE Train local model: \(L_i = \text{Train}(c_i, \text{batch})\)
            \IF{$c_i \in P_c$}
                \STATE Apply poisoning: \(\alpha.\omega \rightarrow \omega_\alpha \)
            \ENDIF
        \ENDFOR
        \STATE \(\omega_G = \text{Agg} (\{\omega_{\alpha},\omega_1, ..., \omega_n\})\)
        \FOR{each client \(c_i \in C\)}
            \STATE Update client model: \(\omega_{c_i} \leftarrow \omega_G\)
        \ENDFOR
    \ENDFOR
\ENDFOR
\end{algorithmic}
\end{algorithm}

\section{Results}

\begin{table}[]
\caption{Number of failed clients to the number of poisoned clients ($p_c$) on different datasets.}
\centering
\begin{tabular}{l|rrr}
\hline
$p_c$& 0    & 0.2   & 0.5   \\ \hline
\textbf{adult}       & 0.00 & 15.79 & 15.79 \\
\textbf{aloi}        & 0.00 & 15.79 & 15.79 \\
\textbf{covtype}     & 0.00 & 15.79 & 15.79 \\
\textbf{epsilon}     & 0.00 & 0.00  & 0.00  \\
\textbf{helena}      & 0.00 & 15.79 & 15.79 \\
\textbf{higgs small} & 0.00 & 15.79 & 15.79 \\
\textbf{jannis}      & 0.00 & 15.79 & 15.79 \\
\textbf{microsoft}   & 0.00 & 15.79 & 15.79 \\
\textbf{yahoo}       & 0.00 & 15.79 & 15.79 \\
\textbf{year}        & 0.00 & 15.79 & 16.45 \\ \hline
\end{tabular}
\label{tab:pc}
\end{table}

Table \ref{tab:pc} shows the number of failed clients to the number of poisoned clients ($p_c$) on different dataset., with percentages representing poisoned clients (0\%, 20\%, and 50\%). Most dataset like "adult," "aloi," and "covtype" show no impact (0.00) when unpoisoned, but a consistent impact (15.79) when 20\% or 50\% clients are poisoned. The "epsilon" dataset remains unaffected (0.00) across all scenarios, while the "year" dataset shows a slightly higher impact (16.45) with 50\% clients poisoned. From the Table, it is clear that CFL is able to survive with low fail rate $(\text{Fail} \leq 16\%)$ in all the experiments.
\begin{table}[]
\caption{Number of failed clients with different data poisoning levels ($p_l$).}
\centering
\begin{tabular}{l|rrrr}
\hline
$p_l$& 0.1   & 0.5   & 1    & 2     \\ \hline
\textbf{adult}       & 10.53 & 10.53 & 0.00 & 10.53 \\
\textbf{aloi}        & 10.53 & 10.53 & 0.00 & 10.53 \\
\textbf{covtype}     & 10.53 & 10.53 & 0.00 & 10.53 \\
\textbf{epsilon}     & 0.00  & 0.00  & 0.00 & 0.00  \\
\textbf{helena}      & 10.53 & 10.53 & 0.00 & 10.53 \\
\textbf{higgs small} & 10.53 & 10.53 & 0.00 & 10.53 \\
\textbf{jannis}      & 10.53 & 10.53 & 0.00 & 10.53 \\
\textbf{microsoft}   & 10.53 & 10.53 & 0.00 & 10.53 \\
\textbf{yahoo}       & 10.53 & 10.53 & 0.00 & 10.53 \\
\textbf{year}        & 11.18 & 10.53 & 0.00 & 10.53 \\ \hline
\end{tabular}
\label{tab:pl}
\end{table}

Table \ref{tab:pl} shows the number of failed clients with varying data poisoning levels $p_l$ (0.1, 0.5, 1, and 2) on different datasets. Most datasets, including adult, aloi, covtype, and others, maintain a value of 10.53 at lower poisoning levels (0.1 and 0.5), drop to 0.00 at level 1, and return to 10.53 at level 2. The epsilon dataset remained at 0.00 across all levels, while the year dataset varied slightly with values of 11.18 at 0.1, 10.53 at 0.5, dropping to 0.00 at level 1, and returning to 10.53 at level 2.  From the Table, it is clear that CFL is able to survive with low fail rate $(\text{Fail} \leq 11\%)$ in all the experiments.

\begin{table}[]
\caption{Number of failed clients compared to the percentage of clients being used for aggregation during Federated Learning ($r_l$) across different datasets.}
\centering
\begin{tabular}{l|rrr}
\hline
$r_l$& 0.2   & 0.8  & 1    \\ \hline
\textbf{adult}       & 31.58 & 0.00 & 0.00 \\
\textbf{aloi}        & 31.58 & 0.00 & 0.00 \\
\textbf{covtype}     & 31.58 & 0.00 & 0.00 \\
\textbf{epsilon}     & 0.00  & 0.00 & 0.00 \\
\textbf{helena}      & 31.58 & 0.00 & 0.00 \\
\textbf{higgs small} & 31.58 & 0.00 & 0.00 \\
\textbf{jannis}      & 31.58 & 0.00 & 0.00 \\
\textbf{microsoft}   & 31.58 & 0.00 & 0.00 \\
\textbf{yahoo}       & 31.58 & 0.00 & 0.00 \\
\textbf{year}        & 31.58 & 0.66 & 0.00 \\ \hline
\end{tabular}
\label{tab:rl}
\end{table}

Table \ref{tab:rl} shows the Number of failed clients when a certain percentage of clients is used for aggregation during Federated Learning ($r_l$) across different datasets. Three $r_l$ values were tested: 0.2, 0.8, and 1.0. Most datasets, including adult, covtype, and yahoo, utilised 31.58\% of clients at $r_l$=0.2, with 0\% at $r_l$=0.8 and 1.0. The epsilon dataset showed 0\% client usage for all $r_l$ values. The year dataset had 31.58\% at RL=0.2, 0.66\% at $r_l$=0.8, and 0\% at $r_l$=1.0. Generally, lower $r_l$ values (0.2) engage more clients than higher values (0.8 and 1.0). CFL models are not effective when only 0.2 of the total client number joins the aggregation on the server. 

\begin{figure}[]
    \centering
    \includegraphics[width=1\linewidth]{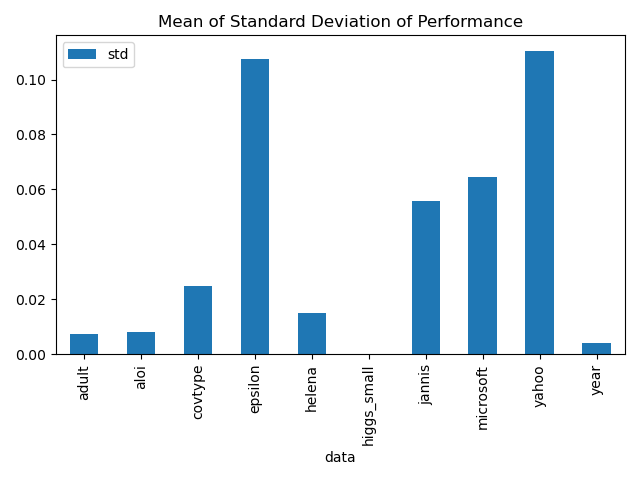}
    \caption{The mean of the standard deviation of the performance across the dataset.}
    \label{fig:meanOfStd}
\end{figure}
Figure \ref{fig:meanOfStd} shows the mean of the standard deviation of the performance across the dataset. Based on the previous result (Table \ref{tab:rl}), this figure was calculated by removing all clients with $r_l = 0.2$. From the figure, we can see that the mean of the standard deviation is small $\mu_{std} \leq 0.1$. The highest mean values are in Epsilon and Yahoo, which have the largest feature sets. From this figure, it is clear that CFL is able to maintain performance across different experiment settings. 

From the theoretical analysis section, our experiment findings can be explained that $r_l$ reduce attack success probability.  In addition, model convergence remains stable when $r_l  \geq 0.8$.

\section{Conclusion}
Random client selection is an effective strategy to defend against adversarial attacks in contrastive federated learning (CFL). Our research demonstrates that CFL can effectively address model poisoning attacks through the use of random client selection in many experiments. This study focuses specifically on model scaling attacks and random selection defence, leaving other attack vectors and defence strategies for future work. Additionally, while our empirical results are promising, we provide only basic theoretical guarantees. 
\section{Acknowledgements}
Leave blank due to anonymity. 

\bibliographystyle{kr}
\bibliography{sample}

\end{document}